\pdfoutput=1

\documentclass[11pt]{article}

\usepackage[]{acl}

\usepackage{times}
\usepackage{latexsym}

\usepackage[T1]{fontenc}

\usepackage[utf8]{inputenc}

\usepackage{microtype}

\usepackage{siunitx}
\usepackage{booktabs}
\usepackage{amsmath}
\usepackage[noend]{algorithm2e}
\usepackage{calc}
\usepackage{array}
\usepackage{graphicx}

\RestyleAlgo{ruled}
\SetKwComment{Comment}{/* }{ */}

\title{UMass PCL at SemEval-2022 Task 4: Pre-trained Language Model Ensembles for Detecting Patronizing and Condescending Language }

\author{David Koleczek \and Alex Scarlatos \and Siddha Karakare \and Preshma Linet Pereira \\
        University of Massachusetts Amherst \\
    \texttt{\{dkoleczek, ajscarlatos, skarkare,  ppereira\}@umass.edu}  \\}

\begin{document}
\maketitle
\begin{abstract}
Patronizing and condescending language (PCL) is everywhere, but rarely is the focus on its use by media towards vulnerable communities. Accurately detecting PCL of this form is a difficult task due to limited labeled data and how subtle it can be. In this paper, we describe our system for detecting such language which was submitted to SemEval 2022 Task 4: Patronizing and Condescending Language Detection. Our approach uses an ensemble of pre-trained language models, data augmentation, and optimizing the threshold for detection. Experimental results on the evaluation dataset released by the competition hosts show that our work is reliably able to detect PCL, achieving an F1 score of 55.47\% on the binary classification task and a macro F1 score of 36.25\% on the fine-grained, multi-label detection task.

\end{abstract}

\section{Introduction}
Everyone has seen, experienced, and expressed patronizing and condescending language (PCL). Someone is patronizing or condescending when they communicate in a way that talks down to others, positions themselves in a superior position to the subjective group, or describes them in a charitable way in order to raise a feeling of compassion for the target person or group. Such language is seen frequently in social media or other mediums where hate is the norm. Many previous works have addressed offensive language and hate speech \cite{zampieri-etal-2019-semeval} \cite{basile2019semeval} in such communities. However, until \cite{perez-almendros2020dontpatronizeme} there has not been work in modeling PCL where 1) it is targeted towards vulnerable communities such as refugees or poor families 2) occurs in the \textit{general media or news}. The authors note that this form of PCL is not usually meant to be harmful, and can often have good intentions, such as when the author is trying to raise awareness about an at-risk group. Yet, research in sociolinguistics has shown that regardless of the intent, PCL has negative effects on potential vulnerable communities such as perpetuating stereotypes, reinforcing societal misbehaviors, and oversimplifying deep-rooted problems. The authors of \textit{Don’t Patronize Me! An Annotated Dataset with Patronizing and Condescending Language towards Vulnerable Communities} originally proposed a new dataset to help in detecting such cases of PCL and are currently hosting a SemEval-2022 competition: Patronizing and Condescending Language Detection \cite{perezalmendros2022semeval}.

In this paper we describe our contribution to both competition subtasks: binary and multi-label classification. Our final ensemble leverages many techniques including pre-trained language models, data augmentation, intermediate fine-tuning on related datasets, and optimizing the detection threshold to empirically maximize F1. For the binary task, our system achieves an F1 score of 55.47\%, or 6.36\% higher than a comparable RoBERTa-based baseline model.
We achieve a macro F1 score of 36.25\% on the multi-label task, or 25.84\% higher than a similar RoBERTa baseline. These results show encouraging progress for the task of detecting fine-grained and linguistically subtle types of PCL, although there is still room for improvement.

\section{Background}
\begin{table*}[h!]
    \centering
    \small
    \begin{tabular}{p{10cm} p{1cm} p{1.2cm} p{1.2cm} }
    
        \textbf{Text} & \textbf{Keyword} & \textbf{Country} & \textbf{Label}\\
        \hline
        
        This wally of an MP just wants his name in the press , what has he ever achieved fro Southend , what did this idle wally do to get the scanner tuned on at the hospital ... zero ! He raised in parliament the fact that his was his mummies birthday , but forgot to mention the homeless families on our streets . & homeless & gb & 3\\
        \hline
        Our life has completely changed from when he as an able-bodied young man running around 5000 miles an hour organising everyone . Now , he 's more disabled than anyone that he ever helped . & disabled & nz & 4\\
        \hline
        The World Health Organization did not give a reason for the increase in deaths , but a provincial health official in Sindh said that the disease hit areas where poor families did not vaccinate their children . & poor-families & us & 2\\
        \hline
        However , she said immigrant patients urgently need treatment and counselling from health-care providers who speak Punjabi or Hindi , and that 's what Roshni -- which means light -- will offer them.  & immigrant & ca & 1\\
        \hline
        
        In February 2015 , five of the men , aged 23 to 25 at the time , went to the station to protest the "" arbitrary and violent "" arrest of one of their friends from the Cova da Moura neighbourhood , known for its large population of immigrants from Cape Verde , a former Portuguese colony off Africa 's northwest coast . & immigrant	& in &	0\\
        \hline

    \end{tabular}
    \caption{Samples from the competition dataset}
    \label{tab:dpm_table}
\end{table*}

\begin{table}[h!]
    \centering
    \small
    \begin{tabular}{l c c c}
        & \textbf{Train} & \textbf{Validation} & \textbf{Test}\\
        \hline
        \textbf{Positive} & 683 & 111 & 199\\
        \hline
        \textbf{Negative} & 6436 & 1145 & 1894\\
        \hline
    \end{tabular}
    \caption{Label distribution in train, validation and test splits for the binary classification subtask}
    \label{tab:label_dist_table}
\end{table}

The Patronizing and Condescending Language Detection task contains two subtasks. In the first subtask, binary classification, the system is given a paragraph of text and it must predict if it contains \textit{any} form of PCL. In the second subtask, multi-label classification, a system is given a paragraph of text and it must identify if it contains each of seven types of PCL (one paragraph can contain multiple categories of PCL).

\subsection{Dataset}
The binary classification task dataset contains 10,636 total labeled text samples that were selected from news stories from the News on the Web corpus \cite{nowcorpus}. All the selected paragraphs are from news stories that were published between 2010 and 2018 from 20 English speaking countries. Each paragraph is assigned a label ranging from 0 (not containing PCL) to 4 (being highly patronizing or condescending). Each paragraph is also associated with one of ten keywords used to retrieve the text from the corpus. Samples from this dataset can be found in Table \ref{tab:dpm_table} and the distribution of positive and negative labels in each of our data splits can be found in Table \ref{tab:label_dist_table}.

The dataset for the multi-label classification task contains paragraphs from the binary classification task which were labeled as containing PCL. Each paragraph is further labeled with at least one of the seven classes: unbalanced power relations, shallow solution, presupposition, authority voice, metaphor, compassion, and the poorer, the merrier. Each is associated with a country-code and keyword as before. Additionally, each paragraph has the start and end spans of where in the paragraph the particular category of PCL is contained. Further background information about this dataset can be found in the paper published by the competition hosts \cite{perez-almendros2020dontpatronizeme}.

\subsection{Baselines}
Recent advances in natural language processing have enabled complex and subtle tasks to be solved by fine-tuning large, pre-trained language models \cite{kalyan2021ammus}. These models can also be fine-tuned on small datasets (such as only 1,000 samples) and while still achieving significant predictive power \cite{mosbach2020stability}, or even with just a few samples \cite{brown2020language}. As such, the competition authors provided baseline results using a pre-trained RoBERTa model, one of the go-to models for fine-tuning for a classification task \cite{liu2019roberta}. They provide results of multiple baselines which help put into context the improvements made by the participating systems. For example, one such baseline is simply random guessing, or assigning each paragraph a random label with equal probability. In our work we build upon the RoBERTa baseline by applying techniques often used in machine learning. Namely, we use early stopping using a validation dataset and set hyperparameters according to best practices discussed by \cite{mosbach2020stability}. We also will experiment with a host of NLP techniques including data augmentation and other language models.

\section{Subtask 1: System Overview}
Considering the challenge of the task and the scarcity of positive examples of PCL, our approach was to experiment with methods focused on enriching our small dataset, such as data augmentation and external datasets. We started with fine-tuning a pre-trained RoBERTa \cite{liu2019roberta} model to use as a foundation. RoBERTa has been shown to be very powerful for a range of NLP tasks such as text classification, while also being straightforward to implement and train using libraries such as HuggingFace \cite{wolf2019huggingface}. We use its architecture unchanged with pretrained weights downloaded from HuggingFace. We add a standard binary classification head on top of RoBERTa's pooled outputs. For preprocessing input into the model, we use HuggingFace's auto-tokenizer. In the following subsections, we detail the additional methods and modifications to the RoBERTa baseline. The models used in our final ensemble use a mix of these methods and the exact ensemble methodology is described in section \ref{sys1ensemble}.

\subsection{Pre-trained Language Models}
In addition to RoBERTa, we use MPNet \cite{song2020mpnet}. This model inherits the pre-training advantages of both Masked Language Modeling \cite{devlin2018bert} and permuted language modeling \cite{yang2019xlnet} under a unified view which splits tokens in a sequence into non-predicted and predicted parts. Experimental results show that MPNet outperforms previous models like BERT, XLNet and RoBERTa by 4.8, 3.4 and 1.5 points respectively on GLUE tasks. In our work, we fine-tune MPNet using the hyperparameters suggested by the authors, which can be found in Appendix \ref{sec:finehyper}.

\subsection{Intermediate Fine-Tuning}

We leverage several external data sources which are related to our task of PCL detection. \cite{phang2018sentence} originally proposed \textit{Supplementary Training on Intermediate Labeled Data Tasks} (STILTs), which we refer to as intermediate fine-tuning. They observed that by 1) starting from a pre-trained language model such as BERT, 2) further training it on an intermediate, \textit{labeled} task, and 3) finally fine-tuned on the target task; improved final performance on a variety of NLP tasks. \cite{pruksachatkun2020intermediate} and \cite{vu2020exploring} provide further insight into tasks that are well suited for intermediate fine-tuning. We used these insights a to develop four independent binary classification tasks on Stereoset \cite{nadeem2020stereoset}, BiasCorp \cite{onabola2021hbert+}, Social Bias Inference Corpus (SBIC) \cite{sap2020socialbiasframes}, and TalkDown \cite{wang2019talkdown} datasets, and used them for intermediate fine-tuning. Two of these tasks, SBIC and Biascorp, are used in our final ensemble, while the other two were left out due to having no performance uplift on our task's test set.

\textbf{BiasCorp} is a dataset containing about 43,000 labeled examples of racial bias from Fox News, BreitbartNews and YouTube. Each text has 3 bias ratings from 0 to 5 and corresponding annotator confidence scores from 1 to 10. To turn this into a binary classification task, we use a similar approach from their paper by first computing a confidence weighted score, and then we threshold this score where scores above 1 are labeled as positive examples.

The \textbf{Social Bias Inference Corpus} is a labeled dataset of over 125,000 examples of social media posts from Reddit, Twitter, and hate sites like Stormfront. For our binary classification task, we use the provided offensiveness rating of a post, and use a threshold of 0.3 for positive examples.

\subsection{Sentence-Level Features}
Transformers are generally able to vastly outperform regression on engineered features. However, in some text labeling tasks, such as essay scoring, it has been shown that engineered features can be used in tandem with Transformer output to improve performance \cite{uto-etal-2020-neural}. This can be achieved simply by concatenating a vector of features $\mathbf{f_n}$ to BERT's CLS vector. An extended classification head will now take the extended vector as input, and similarly project to the final prediction. The classification head, including the added weights, is trained along with the fine-tuning of BERT.

Our set of features was inspired by \cite{uto-etal-2020-neural}, but we excluded the readability metrics because they are not as relevant for our task. Specifically, for text sample $\mathbf{x_n}$, we calculate \textit{the number of words, number of sentences, number of exclamation marks, question marks, and commas, average word length, average sentence length, the number of nouns, verbs, adjectives, and adverbs, and the number of stop words}. Each one of these values is used as a separate dimension of $\mathbf{f_n}$. We leveraged the spaCy Python library \cite{spacy2} to calculate these values. We also included the feature calculation as a data pre-processing step, since calculating the features at run-time would be too costly.

\subsection{Original Labels}
Simply collapsing the 0-4 label in the dataset to a binary value results in a loss of relevant information. To combat this, we translate each label to a \textit{probability} of a sample containing PCL. Specifically, we perform the following mapping: 0 - 0.0, 1 - 0.25, 2 - 0.5, 3 - 0.75, 4 - 1.0. We refer to these probabilities as the \textit{original labels}. We adapt our loss function for this reformulated learning setup by computing the binary cross-entropy loss between the predicted probability and the original label.

\subsection{Backtranslation}
Data augmentation for NLP tasks is becoming increasingly popular with a vast variety of existing techniques. Data augmentation refers to any strategy aimed at increasing the amount of data available for training, by only leveraging the existing training set or domain knowledge about the task \cite{feng2021survey}.
\cite{sennrich2015improving} is a model-based technique for data augmentation where the original text is translated into a desired language and then back to the original language to rephrase it. This process can introduce a different style of sentence for the same meaning. We have used French as our intermediate language for backtranslation on our data. To implement backtranslation for our task, we used the MarianMT model \cite{TiedemannThottingalEAMT2020} provided by HuggingFace to convert between French and English.

\subsection{Ensemble Model} \label{sys1ensemble}
Ensembling is a technique that leverages multiple trained models to improve task performance on the test set. For our final ensemble of three independently trained models, we used a relatively simple approach. Given a set of models $M$ and a test dataset $D$, let $\hat{y}_{ni}$ be $M_i$'s estimate of the probability that text sample $x_n$ is a positive sample. This value is taken directly from $M_i$'s softmax layer output. Each label is then assigned a final positive probability of $\hat{y}_{n} = \frac{1}{|M|}\sum^{|M|}_{i=1}{\hat{y}_{ni}}$. We use this augmented softmax score to make the final prediction for each label in the test set.

\subsection{Choosing an Empirically Optimal Threshold for F1 Score}
The F1 score is the metric used for evaluation in the competition and it is defined as the harmonic mean of precision and recall:
\begin{equation*}
    F_1 = \frac{2 * precision * recall}{precision + recall}
\end{equation*}

Note that the output of our transformer models for binary classification is the softmax of a final linear layer. In order to compute the F1 score, those continuous outputs must be binarized. Typically, that is done via a simple transformation:
\begin{equation}
f(x, threshold) = 
    \begin{cases}
  1, &  x \geq threshold \\
  0, &  x < threshold \\
\end{cases}
\end{equation},
where $x$ is the softmaxed model output and threshold is a value commonly set to be 0.5. Since the model is optimized using cross entropy loss, it is unclear if 0.5 is the best threshold to use in order to maximize F1 score. Since the F1 score is a \textit{harmonic mean}, we would like to balance precision and recall in order to achieve the highest score. Roughly, precision can be increased (and recall decreased) by increasing the threshold, and vice versa by decreasing the threshold.

In order to find the best threshold, we use a simple, brute force algorithm to determine the best threshold on the validation data set. The algorithm takes vectors $y\_true$ and $y\_out$, which are the true binary labels and the normalized model output, respectively. For each model output, we treat it as a \textit{possible threshold} and compute the F1 score using the true labels and $y\_out$ binarized via $f(y\_out, \text{model output})$ (assume $f$ operates on vectors). The algorithm outputs the threshold that achieved the highest F1 score. 

\begin{algorithm}[hbt!]
\caption{Brute-force F1 Threshold}\label{alg:two}
Input: y\_true, y\_out \\
best\_score $\gets$ 0 \\
best\_threshold $\gets$ 0.5 \\
    \For{\upshape threshold in y\_out}{
        bin\_y\_out $\gets$ $f$(y\_out, threshold) \\
        score $\gets$ f1\_score(y\_true, bin\_y\_out) \\
        \If{\upshape score > best\_score}{
            best\_score $\gets$ score\\
            best\_threshold $\gets$ threshold
        }
    }
    \Return best\_threshold
\label{optf1}
\end{algorithm}

\section{Subtask 2: System Overview}
Our system for task 2, multi-label classification, adapts many of the techniques we used in the binary classification task. Similar to our task 1 system, we leverage RoBERTa as a base model for further fine-tuning. However, the target is now a vector of binary labels. Another way to view the problem is as 7 binary classification tasks, one for each category of PCL. Our final ensemble includes models treating the problem in this way, and also training one multi-label classifier directly. We also independently train a binary classification model to generate a prediction for each class using the same model setup from the binary task. In either case, the final ensemble of models was computed identically to the binary classification task by treating each category independently. The sigmoid function was applied to each model output before creating the ensemble.

\subsection{Data Processing}
The dataset for the multi-label classification task contains only paragraphs from the binary classification task which were labeled as containing PCL. This creates a disconnect between the data that would be seen during evaluation (i.e. there are samples without any category of PCL). To handle this discrepancy in data distributions, we take every non-PCL paragraph from the binary classification task dataset and add it to the multiclass dataset with a label vector of all 0s.

\section{Experimental Setup}
We start by splitting our dataset into a train and test set according to the split provided by the competition authors, which is 80\% train and 20\% test data. From the training split, we take 15\% to use as a validation split. Table \ref{tab:label_dist_table} shows the distribution of positive and negative labels within the train, validation and test splits. 

Using our validation split, for each experiment, we perform a hyperparameter sweep by varying the peak learning rate $\in$ \{3e-5, 2e-5, 1e-5, 5e-6\}. We use early stopping where the model is evaluated at the end of each epoch, and if for 10 consecutive epochs performance does not improve, we stop training and save the model from the epoch with the highest valdiation F1 score. For final evaluation on the test set we use this model that achieved the highest F1 score on the validation set. We choose a batch size that maximizes GPU usage, and use the default maximum sequence length of each model. A listing of the full hyperparameters used can be found in Appendix \ref{sec:finehyper}. Random seeds for Python, \texttt{numpy}, and \texttt{PyTorch} were fixed at 221 for all experiments. 

For our final results and submission to the competition, we use the validation split for early stopping and train including the test dataset. All models were trained with a learning rate of 2e-5 unless otherwise noted. We submitted two identical systems, only differentiated by the random seed used. All results shown in Tables \ref{binaryresults} and \ref{mcresults} are the random seed that performed better.

Our experiments were run a system with the following hardware; GPU: NVIDIA RTX 3090, CPU: AMD 5950X, RAM: 64GB 3200Mhz, SSD: 4TB SATA.

\newcommand{\ra}[1]{\renewcommand{\arraystretch}{#1}}
\begin{table}\centering
    \small
    \ra{1.3}
    \begin{tabular}{@{}l p{2.5cm} c c c @{}}
    
    \textbf{Binary} & Description & P & R & F1 \\ 
    \midrule
    Model 1 & RoBERTa, intermediate fine-tuning, features   & 67.95 & 47.75 & 56.08 \\
    Model 2 & MPNet, original labels                        & 65.95 & 55.85 & 60.48 \\
    Model 3 & RoBERTa, features, backtranslation            & 61.11 & 49.55 & 54.73 \\
    \midrule
    
    Ensemble &                              & 62.04 & 60.36 & 61.19 \\
    
    \end{tabular}
    \caption{Subtask 1 -  Validation split scores for each of our models that composed the final ensemble. The scores for each model use the default threshold, while the ensemble's threshold is optimal for the validation split. }
    \label{binaryresults}
\end{table}

\begin{table}\centering
    \small
    \ra{1.3}
    \begin{tabular}{@{}l  p{2.5cm}  c c c @{}}
    
    \textbf{Multi-label} & Description & P & R & F1 \\ 
    \midrule
   
    Model 1 & RoBERTa, features, backtranslation    & 42.90 & 30.52 & 35.01  \\
    Model 2 & RoBERTa, backtranslation, LR=1e-5        & 46.53 & 36.18 & 39.13  \\
    B4MC & Binary model 1 for each category     & 2.51 & 100.0 & 4.83  \\
    \midrule
    
    Ensemble &  & 49.38 & 35.72 & 40.50 \\
    
    \end{tabular}
    \caption{Subtask 2 -  Validation split scores for each of our models that composed the final ensemble.}
    \label{mcresults}
\end{table}

\section{Results}
Submitted systems were evaluated on the F1 score (macro F1 score for the multi-label task). We built several models that each combine different methods. Observe that in Tables \ref{binaryresults} and \ref{mcresults} each model has a varying degree of balance between precision and recall.
This diversity in outcomes and methodology between individual models helps our final ensemble achieve better overall results than any individual model. Our best ensemble for subtask 1 placed 22nd out of 78 and our best ensemble for subtask 2 placed 18th out of 49. 

For subtask 1, our best model on the validation split was the one that used MPNet as a base and used the original labels during training. Model 1 used RoBERTa as a base and was first trained using intermediate fine-tuning, before being trained on the final dataset using sentence level features. Intermediate fine-tuning was done by first training using the SBIC task for 5 epochs, followed by training for 5 epochs on the Biascorp task. Each intermediate task was trained with a learning rate of 5e-6. Finally, model 3 used RoBERTa as a base and was trained with sentence level features and additional backtranslated samples. 

For subtask 2, our best model used RoBERTa as a base and was trained with backtranslated samples and a learning rate of 1e-5. Model 1 was similar except it used sentence level features and our default hyperparameters found in Appendix \ref{sec:finehyper}. Binary for multi-category (B4MC) was 7 individual models (one for each category), trained using the same methodology as our best binary model, model 1.

\subsection{Thresholds}
\begin{figure}
    \centering
    \includegraphics[scale=0.31]{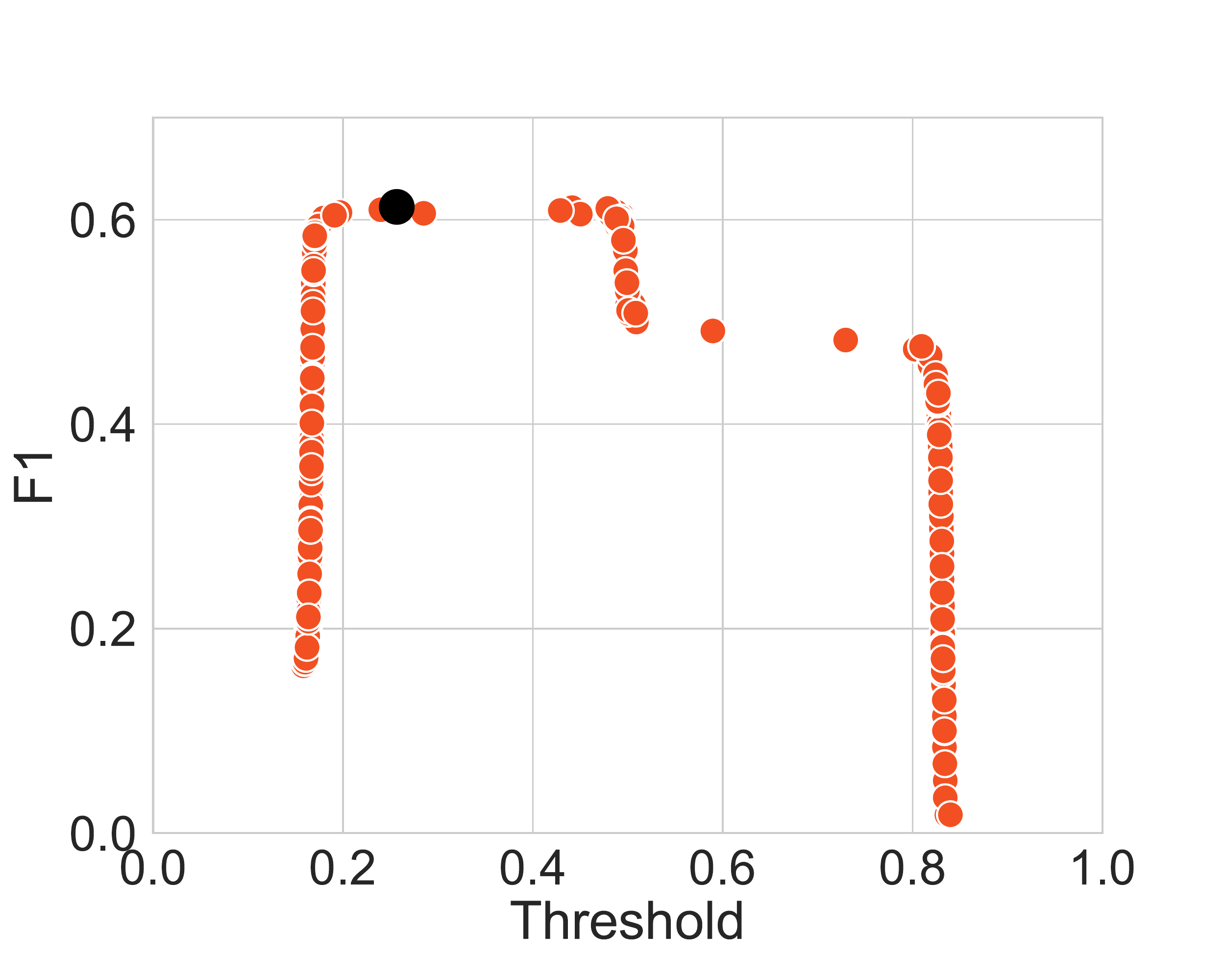}
    \caption{Plot of the resulting F1 score when using each validation softmax score (unbinarized) as the detection threshold.}
    \label{fig:binary_threshold}
\end{figure}

We use Algorithm \ref{optf1} to find an optimal F1 threshold on the validation split in order to increase the F1 score for our final ensemble. Analysis of this optimization for our binary task ensemble is in Figure \ref{fig:binary_threshold}. Based on that plot, we set our prediction threshold to be 0.32, about in the middle of the range of thresholds that achieved the highest F1 scores. Notice that if we had picked a threshold of 0.5, we likely would have achieved a lower score due to the drop in scores around that mark. 

We performed similar analysis for subtask 2. The thresholds we choose were 0.5, 0.31, 0.49, 0.27, 0.29, 0.21, and 0.4 for the categories unbalanced power relations, shallow solution, presupposition, authority voice, metaphor, compassion, and the poorer the merrier, respectively.

\subsection{Error Analysis}

\begin{table*}[h!]
    \centering
    \small
    \begin{tabular}{p{7cm} p{1cm} p{1.2cm} p{1.2cm} p{2cm}}
        \textbf{Text} & \textbf{Label} & \textbf{Baseline Pred.} & \textbf{Ensemble Pred.} & \textbf{Categories}\\
        \hline
        real poverty of britain : shocking images of uk in the sixties where poor really meant poor these hard-hitting photographs offer a glimpse into the harrowing day-to-day for poor families living in britain during the sixties . & pos. & neg. & pos. & Authority Voice, Compassion\\
        \hline
        rather sad . good set of pictures that tells a tale of survival , subsistence living , and hopeless nature of life in the tribal societies . exploiting an unexpected geo-political bonanza is a temporary relief that is not sustainable . education and economic development seem miles away . & pos. & neg. & pos. & Presupposition, Metaphors,	Compassion\\
        \hline
        as a family , my father was a policeman and he used to come home with food ( monthly ) , and my mother used to pack small parcels and we used to give them to the poor families . & pos. & neg. & pos. & Unb. power rel., Shallow solu.\\
        \hline
        mrs charo said some people were injured while escaping the floods . she added that they are now in need of food , clothing and clean drinking water . & neg. & pos. & neg. & -\\
        \hline
        she added : i would also like to carefully point out that the issue was not her religious beliefs , but rather it is about choosing to treat men and women differently by shaking the hands of women but not men . & neg. & pos. & neg. & -\\
        \hline
        i realised that it was not impossible to achieve anything in the world despite being disabled . & neg. & pos. & neg. & -\\
    \end{tabular}
    \caption{Samples where the ensemble improved over the baseline.}
    \label{tab:error_analysis_ours_right}
\end{table*}

\begin{table*}[h!]
    \centering
    \small
    \begin{tabular}{p{7cm} p{1cm} p{1.2cm} p{1.2cm} p{2cm}}
        \textbf{Text} & \textbf{Label} & \textbf{Baseline Pred.} & \textbf{Ensemble Pred.} & \textbf{Categories}\\
        \hline
        many refugees do n't want to be resettled anywhere , let alone in the us . & pos. & neg. & neg. & Presupposition \\
        \hline
        the law stipulated 21 rights of the disabled persons . the disabled persons must get the national identity cards and be listed in the voters roll . even they will be able to contest in the polls . & pos. & neg. & neg. & Unb. power rel.\\
        \hline
        however , this success story is not uncommon . it often happens that people from poor families have more perseverance to fight tooth an nail in business than children of rich parents who are used to get everything they want with ease . people without a strong spirit will quickly break down and drop out from the competition . & pos. & neg. & neg.  & Presupposition, Metaphors, The p., the mer.\\
        \hline
        your personal leadership has been critical to addressing the plight of the rohingya who fled to safety in your country . i thank you for all you have done to assist these men , women and children in need , he wrote in the message . & neg. & pos. & pos. & -\\
        \hline
        she also praised the efforts to stabilise the lives of refugees , by providing for their basic needs and helping them overcome their challenges , while stressing that supporting refugees is an ongoing part of the uae 's humanitarian directives , and the country has taken the responsibility to evaluate their needs and provide them with a variety of urgent and essential aid . & neg. & pos. & pos. & -\\
    \end{tabular}
    \caption{Samples where both the baseline and ensemble were incorrect.}
    \label{tab:error_analysis_both_wrong}
\end{table*}

To further analyze how our model compares to the baseline, we examine samples from the test set where 1) the baseline made an error, but our model made the correct prediction, and 2) where both the baseline and our model made the incorrect prediction. We examine false positives and false negatives in both cases.

In Table \ref{tab:error_analysis_ours_right}, we can see that the baseline's false negatives tend to be more poetic examples. They paint vivid, dramatic scenes, and use strong adjectives like "shocking" and "hopeless". This indicates that our model may have picked up on this poetic language signal as an indicator of PCL.

In the baseline's false positives, we see phrases that may be indicative of PCL, such as "in need of food", "treat men and women differently", and "being disabled". However, in context, these phrases are not condescending to the subjects of the statements. It's possible that the baseline was simply acting on the presence of these phrases, while our model was able to identify the lack of other supporting signals as evidence that the sentences were not PCL.

In Table \ref{tab:error_analysis_both_wrong}, we can see where both the baseline and our model failed, where there are a variety of possible issues. The first two false negatives are only identifiable as PCL given their subtext (assuming the perspective of refugees and creating a division between disabled and non-disabled people), rather than obvious choices of words. While the third false positive uses flowery language (a possible PCL indicator), the identifying piece, "people from poor families have more perseverence...", uses a positive sentiment. The models seem to struggle with the "poorer the merrier" category, which this is an example of. It's possible that the model is overwhelmed with more negative sentiments from other categories so it tends to correlate that too strongly with PCL.

The false positives all use the kinds of words that seem to be common in the actual positive examples, such as "plight", "challenges", "disadvantaged", and "vulnerable". What the models seem to fail at is distinguishing between pointing out a group's disadvantage and using that disadvantage to patronize the group.

Overall, the models seem to be very active on words and phrases that might indicate PCL. While our best model appears able to deal with these features in somewhat intelligent ways, the main downside is that it cannot place the samples in a global context. Simply using certain adjectives to describe a certain group of people does not necessarily make a statement condescending. It depends on exactly which adjectives, which group, and what other context is in the statement. As humans, we have two main advantages in this task: 1) we understand the struggles and stereotypes of disadvantaged groups, and 2) we also can perform deductive reasoning to determine the underlying meaning of a phrase. Possibly remedies for our model might be 1) providing many more examples, so it can build the kind of context needed to make these distinctions, and 2) to enhance the model with ways of reasoning about the problem.

\section{Conclusion}
We developed a system to detect patronizing and condescending language by media towards vulnerable communities. We combined a multitude of NLP techniques in our final ensemble. Our system leveraged backtranslation, sentence level features, intermediate fine-tuning, and pre-trained language models. We also proposed multiple training techniques and optimizing the F1 threshold. These improvements resulted in our system placing in the top 25\% for subtask 1 and the top 40\% for subtask 2.

\section*{Acknowledgements}

This work began as part of an advanced natural language processing course at the University of Massachusetts Amherst. We thank the teaching assistants Katherine Thai and Tu Vu and the instructor Mohit Iyyer for their wonderful teaching and feedback.

\bibliography{anthology,custom}
\bibliographystyle{acl_natbib}

\appendix

\section{Fine-tuning Hyperparameters} \label{sec:finehyper}
\begin{table}[!h]
    \centering
    \begin{tabular}{l|c c c}
    \toprule
     \textbf{Hyper-parameter} & Fine-tuning \\
     \midrule
        Learning Rate & 2e-5 \\
        Batch Size & 16 \\
        Weight Decay & 0.1 \\
        Epochs & Early stopping \\
        Learning Rate Decay & Linear \\
        Warmup Ratio & 0.06 \\
    \bottomrule
    \end{tabular}
    \caption{Hyper-parameters for fine-tuning RoBERTa and MPNet models.}
    \label{fine_tuning_params}
\end{table}

\end{document}